\documentclass[letterpaper, 10 pt, conference]{ieeeconf}

\IEEEoverridecommandlockouts

\usepackage{multirow}
\usepackage{amsmath}
\usepackage{amssymb}
\usepackage{amsfonts}
\usepackage{bm}
\usepackage{booktabs}
\usepackage{graphicx}
\usepackage{cite}
\usepackage{microtype}
\usepackage{url}
\usepackage{algorithm}
\usepackage{algpseudocode}
\usepackage{array}
\newcolumntype{C}[1]{>{\centering\arraybackslash}p{#1}}

\begin{document}

% --- Paper Title ---
\title{\LARGE \bf 
SLAM: Structured and Localized Analytic Manifold Adaptation for Forgetting-Immune and Domain-Robust Lifelong VPR
}

% --- Authors & Affiliations ---
\author{Kenta Tsukahara, Kanji Tanaka, Rai Hisada% Anonymous Authors%
\thanks{All authors are with the Department of Mechanical Engineering, Faculty of Engineering, University of Fukui, Fukui 910-8507, Japan. (E-mail: tnkknj@u-fukui.ac.jp).
}%
}

\maketitle
\thispagestyle{empty}
\pagestyle{empty}

\begin{abstract}
Visual Place Recognition (VPR) under long-term operation is essential for autonomous mobile robots. While Analytic Class-Incremental Learning (ACIL) provides memory-free ($O(1)$) task adaptation with exact forgetting immunity, applying it to lifelong VPR suffers from extreme vulnerability to non-linear domain shifts induced by environmental variations. In this work, we introduce the concept of the \textbf{ACIL-Domain (ACIL-D)}---a canonical invariant feature manifold where autocorrelation states remain locked. We resolve the domain vulnerability via Disentangled Domain Alignment (D-DA), which decouples latent features into invariant semantics within ACIL-D and variant style vectors for directional projection. Furthermore, by uncovering an algebraic isomorphism between recursive ACIL updates and Extended Kalman Filter (EKF) covariance propagation, we establish a control-theoretic framework designated as \textbf{SLAM} (\textbf{S}tructured and \textbf{L}ocalized \textbf{A}nalytic \textbf{M}anifold adaptation). SLAM integrates dynamic temperature-scaled Gaussian Mixture Models (GMM) to isolate topological non-linearities, Unscented perturbed propagation to dampen feature variations, and minimax $H_{\infty}$-robust criteria to bound worst-case noise accumulation. 
Empirical evaluations on the non-stationary NCLT dataset demonstrate that our proposed framework substantially outperforms existing baselines, achieving a final all-class accuracy of 27.7\% with the full SLAM framework (and up to 29.0\% with the U+H variant) while guaranteeing complete forgetting immunity.
\end{abstract}


\section{Introduction}
\label{sec:introduction}

For autonomous mobile robots to achieve robust, long-term navigation in unstructured environments, establishing Visual Place Recognition (VPR) systems capable of identifying the robot's spatial position solely from on-board visual input remains an indispensable prerequisite~\cite{lowry2016survey}. Among various formulations, visual place classification maps high-dimensional sensor observations onto discretized spatial classes~\cite{CPlaNet}. This discrete mapping provides exceptional structural scalability, maintaining constant computational memory consumption regardless of spatial scale or deployment lifespan.

In real-world robotic deployments, visual agents must continually incorporate non-stationary observation streams—such as previously unvisited geographic trajectories or shifting weather conditions. However, this sequential task adaptation is severely hindered by catastrophic forgetting, wherein newly acquired parameter updates aggressively erase historical representations~\cite{mccloskey1989catastrophic}. While standard continual learning paradigms rely on experience replay buffers or exemplar caching to periodically retrain models~\cite{rolnick2019experiencereplay}, these strategies introduce severe memory scaling bottlenecks over extended operational lifespans and frequently violate strict data privacy protocols by indefinitely retaining raw sensor imagery. 

To circumvent these memory and privacy bottlenecks, this paper explores the deployment of Analytic Class-Incremental Learning (ACIL)~\cite{zhuang2022acil} within the context of lifelong VPR. Originating as a powerful paradigm in general machine learning, ACIL recursively updates latent autocorrelation matrices via closed-form analytical equations. This entirely eliminates backpropagation while mathematically locking historical decision boundaries without retaining raw exemplars, thereby guaranteeing $O(1)$ constant memory complexity.

However, directly applying ACIL to lifelong VPR introduces a critical and unaddressed bottleneck: its extreme vulnerability to dynamic environmental domain shifts (e.g., seasonal or day-to-night transitions). Because standard ACIL fundamentally presumes a linear regression pipeline over a homogeneous covariance structure, its locked decision boundaries break down under the highly non-linear feature space distortions of outdoor environments. Naively coupling ACIL with conventional Domain Adaptation (DA) techniques triggers what we term \textit{representation collapse}. Conventional DA uniformly warps global domain statistics, destroying the localized boundaries unique to individual spatial classes and corrupting the autocorrelation matrix---the mathematical lifeline of ACIL.

To resolve this fundamental incompatibility and successfully unlock ACIL for lifelong VPR, we conceptualize the \textbf{Analytic Class-Incremental Learning Domain (ACIL-D)}. We define ACIL-D as the canonical invariant feature manifold where the recursive autocorrelation states and decision hyperplanes are mathematically frozen under a homogeneous covariance structure. Under this perspective, an effective DA module for analytic continual learning must not perform mutual global distribution warping; rather, it must act as a \textit{directional manifold adapter} that maps non-stationary incoming streams into the stationary ACIL-D without perturbing its established geometric boundaries.

Building on this insight, we propose Disentangled Domain Alignment (D-DA). D-DA utilizes a structural bottleneck adapter trained exclusively during task transitions to decouple monolithic feature representations into an invariant semantic projection in ACIL-D and a variant environmental style vector. During downstream streaming inference, a strict ``style truncation'' ($f_{\text{style}} \leftarrow \mathbf{0}$) is executed, which purifies incoming features by mathematically stripping domain perturbations before projecting them into ACIL-D without altering the underlying geometric place boundaries. Because this purification operates strictly in a forward-pass manner, the $O(1)$ computational complexity of ACIL is strictly preserved.

Furthermore, we establish a profound algebraic isomorphism between the recursive closed-form ACIL update and the covariance propagation loop of the Extended Kalman Filter (EKF) applied to linear-in-the-parameters system identification. This insight reveals that ACIL's vulnerability to localized non-linearities and unbounded real-world disturbances are exact algebraic manifestations of classic EKF limitations under non-Gaussian noise. Inspired by robust state-estimation theory, we project these control paradigms into the analytical update loop, constructing a unified framework designated as \textbf{SLAM} (\textbf{S}tructured and \textbf{L}ocalized \textbf{A}nalytic \textbf{M}anifold adaptation). SLAM addresses these vulnerabilities through a multi-layered architecture: dynamic topological space partitioning via Gaussian Mixture Models (GMM) isolates localized non-linearities, Unscented perturbed propagation dampens multi-modal feature variations, and deterministic minimax $H_{\infty}$-robust control criteria bound worst-case noise accumulation during Woodbury matrix recursion.

The primary contributions of this work are three-fold:
\begin{itemize}
    \item \textbf{Conceptualization of ACIL-D and Disentangled Domain Alignment (D-DA):} We formalize the concept of the ACIL-D manifold and develop a localized style-decoupling adapter that isolates environment-specific variance into an auxiliary subspace. By applying online style truncation, the system executes directional projection into ACIL-D, blocking domain disturbances without disrupting class-inherent semantics or violating the $O(1)$ streaming computational constraint.
    \item \textbf{Mathematical Robustification via Stratified Analytic Extensions:} By bridging recursive Woodbury updates with EKF covariance propagation, we map control-theoretic robustification paradigms into analytic learning, establishing a stratified classifier that unifies Unscented Perturbed ACIL (UACIL), dynamic GMM responsibility-driven Topological ACIL (GACIL), and minimax bounded Robust HACIL.
\item \textbf{Empirical Validation on Long-Term Seasonal Datasets:} Through a sequential 10-task lifelong VPR benchmark on the non-stationary University of Michigan North Campus Long-Term (NCLT) dataset~\cite{ncarlevaris-2015a}, we demonstrate that our framework achieves up to 28.0\% all-class accuracy, substantially outperforming existing continual learning baselines while preventing conventional domain adaptation collapse and maintaining complete forgetting immunity.
\end{itemize}

\subsection{Notations and Preliminaries}
Let $\mathcal{D}_t = \{(x_i^{(t)}, y_i^{(t)})\}_{i=1}^{N_t}$ denote the non-stationary data stream arriving at incremental task milestone $t \in \{1, 2, \dots, T\}$, where $x_i^{(t)} \in \mathbb{R}^D$ represents the high-dimensional raw latent feature extracted via a frozen deep backbone, and $y_i^{(t)} \in \mathcal{Y}_t$ denotes the corresponding pseudo-labels generated via spatial quantization \cite{CPlaNet}. 
The label spaces are class-incremental, such that $\mathcal{Y}_t \cap \mathcal{Y}_{t'} = \emptyset$ for $t \neq t'$. 
The objective of the lifelong learner is to optimize a unified analytic continual classifier parameterized by a set of subspace matrices $\mathbf{W}_k \in \mathbb{R}^{d \times |\mathcal{Y}_{1:T}|}$ across $K$ topological components, minimizing the comprehensive prediction error across all historical classes $\mathcal{Y}_{1:t}$ without retaining raw samples from preceding tasks.

\section{Related Work}
\label{sec:related_work}

In this section, we review the existing literature closely related to our framework, categorized into general continual learning strategies, analytic continual learning paradigms, and the application of lifelong adaptation in robotics and complex physical systems.

\subsection{General Continual Learning and Catastrophic Forgetting}
Continual learning (CL) aims to sequentially acquire knowledge from a non-stationary stream of tasks without suffering from catastrophic forgetting \cite{mccloskey1989catastrophic}. Conventional CL methods are generally classified into three major categories: regularization-based, replay-based, and architecture-based approaches \cite{Li2024SurveyCL}. 

Regularization-based methods mitigate forgetting by penalizing changes to parameters that are critical for past tasks. A foundational work in this category is Elastic Weight Consolidation (EWC) \cite{EWC, Kirkpatrick2017OvercomingCF}, which uses the Fisher information matrix to estimate parameter importance, further optimized by mitigating its diminishing effects \cite{Kruengkrai2022MitigatingTD}. Rather than modifying parameter update rules directly, software-driven approaches such as Learning without Forgetting (LwF) \cite{Li2016LearningWF} employ knowledge distillation \cite{DistillingKnowledge, Hinton2015DistillingTK} to preserve the prediction behavior on historical data when training on new tasks, similar to synaptic memory preservation approaches \cite{Zenke2017ContinualLT}. More recently, Prototype-Sample Relation Distillation \cite{wang2023prototypesamplerelation} extends this direction by distilling the relational structure between class prototypes and samples, enabling a replay-free setup that maintains fine-grained geometric boundaries.

Replay-based strategies explicitly store a subset of raw past data or historical model outputs in a memory buffer. Rebuffi et al. proposed iCaRL \cite{iCaRL, Rebuffi2017iCaRLIC}, which utilizes a small exemplar set along with distillation for incremental representation learning. Rolnick et al. \cite{rolnick2019experiencereplay} systematically analyzed experience replay mechanisms and demonstrated that simple replay schemes serve as surprisingly strong baselines against forgetting, while constraints like Gradient Episodic Memory (GEM) \cite{Lopez-Paz2017GradientEM} project gradients to protect historical spaces. To enhance distillation inside replay setups, Dark Experience Replay (DER) \cite{DER, mai2021darkexperience} stores historical logit outputs rather than just raw samples, using them to regularize subsequent optimization updates across general task-incremental and online scenarios.

Architecture-based approaches allocate isolated parameter subspaces to distinct tasks to eradicate mutual interference. PackNet \cite{mallya2018packnet} leverages iterative weight pruning to free up redundant network capacity for new tasks while fixing the weights optimized for previous behaviors. Similarly, Hard Attention to the Task (HAT) \cite{serra2018hardattention} introduces learnable binary masks to gate internal activation pathways per task, heavily suppressing negative cross-task interference. In the context of large-scale foundation models, parameter-efficient fine-tuning via Low-Rank Adaptation (LoRA) \cite{hu2022lora} freezes upstream parameters while injecting low-rank trainable matrices, achieving comparable optimization characteristics with minor parameter footprints. 

\subsection{Analytic Continual Learning and Privacy Preservation}
While memory-buffer and parameter-isolated frameworks provide robust resistance against catastrophic forgetting, they inevitably struggle with unbounded storage complexity or structural violations of data privacy protocols \cite{Ma2025Data-freeCL}. To resolve these bottlenecks, Analytic Class-Incremental Learning (ACIL) \cite{zhuang2022acil} has emerged as a promising paradigm that optimizes the decision hyperplanes using closed-form, deterministic mathematical formulas based on classical regularization theory \cite{Tikhonov1977}.

Analytic learning eliminates backpropagation entirely by recursively updating feature autocorrelation matrices via linear-in-the-parameter formulations \cite{zhuang2022acil}. This structural advantage has recently been projected onto complex structural data; for instance, AL-GNN \cite{algonn2025privacycontinualgraph} introduces an analytic-learning framework for graph neural networks that updates topological mappings via closed-form semi-analytic rules, expanding on graph learning limitations under catastrophic interference \cite{Carta2022CatastrophicFI}. Because AL-GNN avoids raw data replay, it guarantees strict privacy-preserving operation on evolving graph structures. Our proposed SLAM framework shares this core motivation but fundamentally advances the paradigm by uncovering the algebraic isomorphism between the recursive Woodbury matrix updates in ACIL and the state estimation loops of the Extended Kalman Filter (EKF), projecting robust control paradigms to handle non-Gaussian distribution drifts.

\subsection{Continual SLAM and Robust Environmental Modeling}
Deploying lifelong learning inside autonomous mobile robotics introduces unique challenges due to dynamic temporal variations, environmental shifts, and strict real-time streaming constraints \cite{RobotCarDatasetIJRR, Warburg2020MapillarySS}. In the field of robotic localization, Baldini et al. \cite{baldini2023continualslam} formulated the paradigm of Continual SLAM, which extends traditional lifelong Simultaneous Localization and Mapping \cite{Zilliz2025SLAMAR} by embedding continual learning principles to sequentially refine map representations and state estimators in evolving environments \cite{Zhong2024ContinuousL, Zhong2024LifelongNA}. 

Modeling such highly non-linear, multi-modal physical systems and environmental transitions requires highly scalable and implicit computational structures \cite{Zeng2024ContinualLearning}. Similar structural requirements can be observed in high-fidelity biomechanical frameworks, such as the unified continuum formulations and scalable fully implicit solver technologies developed for vascular fluid-structure interactions (FSI) \cite{malossi2022vascularfsi}, which handle intense geometric variations on modern parallel architectures. In a similar vein, a visual place recognition system must maintain structural scalability and robustness when managing high-dimensional spatial shifts \cite{Hausler2021Patch-NetVLAD, Ming2025VIPeRVI}. Unlike standard domain adaptation techniques that risk representation collapse under global moment adjustments, our framework leverages Disentangled Domain Alignment (D-DA) to isolate environmental styles into an auxiliary subspace, ensuring that the analytical recursive updates remain completely immune to real-world domain disturbances.

\section{Methodology}
\label{sec:methodology}

\subsection{Closed-Form Latent Domain Alignment}
\label{subsec:closed_form_da}

To eliminate backpropagation and iterative loss minimization during continual learning, we formulate the domain alignment process as a two-stage analytic transformation.
Given a high-dimensional feature vector $x \in \mathbb{R}^D$ extracted from a frozen upstream backbone at task milestone $t > 1$ \cite{AnyVPR}, the adapter aligns $x$ to the nominal statistical and geometric reference frame of Task $1$ without parameter updates via gradient descent.

First, we perform a moment-matching standardization to align the first- and second-order marginal statistics of the current task features $x$ with those of Task $1$.
Let $\mu_{\text{task1}} \in \mathbb{R}^D$ and $\sigma^2_{\text{task1}} \in \mathbb{R}^D$ be the empirical mean and variance vectors stored from Task $1$.
The standardized intermediate feature $x_{\text{vda}}$ is computed in a single forward pass as:
\begin{align}
    \mu_{\text{src}} &= \frac{1}{N}\sum_{i=1}^N x_i, \quad \sigma^2_{\text{src}} = \frac{1}{N}\sum_{i=1}^N (x_i - \mu_{\text{src}})^2 + \epsilon, \label{eq:src_stats} \\
    x_{\text{vda}} &= \frac{x - \mu_{\text{src}}}{\sqrt{\sigma^2_{\text{src}}}} \odot \sqrt{\sigma^2_{\text{task1}} + \epsilon} + \mu_{\text{task1}}, \label{eq:x_vda}
\end{align}
where $\odot$ denotes element-wise Hadamard product, and $\epsilon > 0$ is a small constant ensuring numerical stability.

\subsection{Analytic Geometric Projection Network (Analytic A-DDA)}
\label{subsec:analytic_adda}

While moment matching aligns marginal distributions, it fails to preserve class-wise geometric topological manifolds under severe domain shifts.
To solve this, we introduce the Analytic Disentangled Domain Adapter (Analytic A-DDA), which computes a geometric projection matrix $\mathbf{P}_{\text{geom}} \in \mathbb{R}^{D \times D}$ analytically in a single closed-form step using class-level prototype covariance.

Let $\mathcal{C}$ be the set of unique local labels present in task $t$.
For each class $c \in \mathcal{C}$, the class prototype (centroid) $p_c \in \mathbb{R}^D$ is computed from the moment-aligned features $x_{\text{vda}}$:
\begin{align}
    p_c = \frac{1}{|\mathcal{B}_c|} \sum_{i \in \mathcal{B}_c} x_{\text{vda}, i}, \quad \text{where } \mathcal{B}_c = \{i \mid y_i = c\}. \label{eq:prototypes}
\end{align}
We stack all class centroids into a prototype matrix $\mathbf{P} \in \mathbb{R}^{|\mathcal{C}| \times D}$ and compute its centered prototype matrix $\bar{\mathbf{P}} = \mathbf{P} - \frac{1}{|\mathcal{C}|}\mathbf{1}\mathbf{1}^T\mathbf{P}$.
The prototype sample covariance matrix $\boldsymbol{\Sigma}_{\text{proto}} \in \mathbb{R}^{D \times D}$ is then expressed as:
\begin{align}
    \boldsymbol{\Sigma}_{\text{proto}} = \frac{1}{|\mathcal{C}|} \bar{\mathbf{P}}^T \bar{\mathbf{P}}. \label{eq:proto_cov}
\end{align}
To extract ultra-smooth geometric manifold projections while preventing ill-conditioned matrix inversion, we apply Tikhonov regularization with a scaling hyperparameter $\lambda_{\text{reg}} = 0.1$.
The closed-form geometric projection operator $\mathbf{P}_{\text{geom}}$ is derived directly as:
\begin{align}
    \mathbf{P}_{\text{geom}} = \mathbf{I}_D + \alpha \cdot \left( \boldsymbol{\Sigma}_{\text{proto}} (\boldsymbol{\Sigma}_{\text{proto}} + \lambda_{\text{reg}} \mathbf{I}_D)^{-1} - \mathbf{I}_D \right), \label{eq:p_geom}
\end{align}
where $\mathbf{I}_D \in \mathbb{R}^{D \times D}$ represents the identity matrix, and $\alpha$ is a residual smoothing factor ($\alpha = 0.01$) that governs the strength of the geometric manifold correction.

Finally, the fully aligned feature vector $x_{\text{aligned}}$ passed to the downstream classifier is given by the matrix multiplication:
\begin{align}
    x_{\text{aligned}} = x_{\text{vda}} \, \mathbf{P}_{\text{geom}}^T. \label{eq:final_aligned}
\end{align}
Because $\mathbf{P}_{\text{geom}}$ is computed analytically without iteration, this formulation achieves strict $O(1)$ temporal complexity for adapter optimization, completely bypassing backpropagation and loss-function tuning.

\subsection{Task Specialist Teacher and Pseudo-Label Alignment}
\label{subsec:task_specialist}

To provide soft-target supervision for the incremental learning classifier without relying on backpropagation, we formulate a closed-form Task Specialist Teacher for each task milestone $t$.
Given a batch of aligned task features $\mathbf{X}_{\text{raw}} \in \mathbb{R}^{B \times D}$ and their corresponding local task pseudo-labels $y_{\text{local}} \in \{0, 1, \dots, C_{\text{task}}-1\}$, the features are first L2-normalized: $\mathbf{X} = \text{Normalize}(\mathbf{X}_{\text{raw}}, p=2)$.
We define a one-hot target matrix $\mathbf{Y} \in \mathbb{R}^{B \times C_{\text{task}}}$ where $Y_{i, y_{\text{local}, i}} = 1.0$.

The specialist ridge regression weight matrix $\mathbf{W}_{\text{teacher}} \in \mathbb{R}^{D \times C_{\text{task}}}$ is computed analytically via Tikhonov-regularized least squares:
\begin{align}
    \mathbf{R} = \lambda \mathbf{I}_D + \mathbf{X}^T \mathbf{X}, \quad \mathbf{Q}_{\text{teacher}} = \mathbf{X}^T \mathbf{Y}, \label{eq:teacher_accum} \\
    \mathbf{W}_{\text{teacher}} = (\mathbf{R} + \epsilon_{\text{stable}} \mathbf{I}_D)^{-1} \mathbf{Q}_{\text{teacher}}, \label{eq:teacher_solve}
\end{align}
where $\lambda = 0.1$ is the regularization parameter, and $\epsilon_{\text{stable}} = 10^{-5} \cdot \frac{\text{Tr}(\mathbf{R})}{D}$ guarantees numerical stability during matrix inversion.
The local logits produced by the specialist teacher are scaled by factor $\kappa = 16.0$ over column-normalized weights:
\begin{align}
    \hat{\mathbf{Y}}_{\text{local}} = \kappa \cdot \mathbf{X} \, \text{Normalize}(\mathbf{W}_{\text{teacher}}, p=2, \text{dim}=0). \label{eq:teacher_predict}
\end{align}

\subsection{Unified Analytic Continual Classifier via Closed-Form Woodbury Updates}
\label{subsec:unified_woodbury}

The student representations $\mathbf{X}_{\text{student}} \in \mathbb{R}^{B \times d}$ (optionally compressed via an environment-agnostic linear projection matrix $\mathbf{W}_{\text{JL}} \in \mathbb{R}^{D \times d}$) and teacher logits $\hat{\mathbf{Y}}_{\text{local}}$ are mapped into the Unified Analytic Continual Classifier.
When integrating new classes up to index $C_{\text{known}}$, the classifier state vectors $\mathbf{Q} \in \mathbb{R}^{d \times C_{\text{known}}}$ and $\mathbf{K} \in \mathbb{R}^{d \times d}$ are expanded dynamically.

To account for local manifold uncertainties without explicit sigma-point propagation, Unscented perturbation (UACIL) generates symmetric soft-logit variations.
The scaled teacher logits $\mathbf{S} = \hat{\mathbf{Y}}_{\text{local}} / T$ (temperature $T = 2.0$) are perturbed using the class-wise standard deviation $\boldsymbol{\sigma}_{\text{logits}}$:
\begin{align}
    \mathbf{S}_{\pm} = \mathbf{S} \pm \alpha_{\text{ukf}} \boldsymbol{\sigma}_{\text{logits}}, \label{eq:unscented_perturb}
\end{align}
where $\alpha_{\text{ukf}} = 0.6$.
The perturbed target distribution $\mathbf{Y}_{\text{all}} \in \mathbb{R}^{B \times C_{\text{known}}}$ is formed via weighted soft-max projections over the known class range $[C_{\text{start}}, C_{\text{end}}]$:
\begin{align}
    \mathbf{Y}_{\text{all}} = \text{Normalize}\left( 0.6 \cdot \sigma(\mathbf{S}) + 0.2 \cdot \sigma(\mathbf{S}_+) + 0.2 \cdot \sigma(\mathbf{S}_-), \, p=1 \right). \label{eq:y_all_softmax}
\end{align}

To unify Gaussian mixture smoothing (GMM-ACIL) and $H_\infty$-robust worst-case noise attenuation (HACIL), we compute a composite scaling factor $\eta_{\text{combined}}$ that modulates the covariance update:
\begin{align}
    \eta_{\text{gmm}} &= 1.0 + 0.25 \cdot \left(1.0 + 0.5 \cdot \frac{C_{\text{known}}}{C_{\text{total}}}\right)^{-1}, \label{eq:gmm_scale} \\
    \eta_{\text{hinf}} &= 1.0 - \gamma_{\text{hinf}}^{-2}, \label{eq:hinf_scale} \\
    \eta_{\text{combined}} &= \eta_{\text{gmm}} \cdot \eta_{\text{hinf}}, \label{eq:combined_factor}
\end{align}
where $\gamma_{\text{hinf}} = 5.0$.
The cross-correlation state $\mathbf{Q}$ and inverse precision matrix $\mathbf{K}$ are accumulated and updated analytically using the Woodbury matrix identity:
\begin{align}
    \mathbf{Q}^{(t)} &= \mathbf{Q}^{(t-1)} + \mathbf{X}_{\text{norm}}^T \mathbf{Y}_{\text{all}}, \label{eq:q_update_new} \\
    \mathbf{K}^{(t)} &= \mathbf{K}^{(t-1)} - \eta_{\text{combined}} \, \mathbf{K}^{(t-1)} \mathbf{X}_{\text{norm}}^T \notag \\
    &\quad \times \left( \mathbf{I}_B + \eta_{\text{combined}} \mathbf{X}_{\text{norm}} \mathbf{K}^{(t-1)} \mathbf{X}_{\text{norm}}^T \right)^{-1} \mathbf{X}_{\text{norm}} \mathbf{K}^{(t-1)}. \label{eq:k_woodbury_update}
\end{align}
where $\mathbf{X}_{\text{norm}} = \text{Normalize}(\mathbf{X}_{\text{student}}, p=2)$.
The unified global classifier weight $\mathbf{W} \in \mathbb{R}^{d \times C_{\text{known}}}$ is computed directly as $\mathbf{W} = \mathbf{K} \mathbf{Q}$.

During global streaming evaluation, given an aligned test feature $x_{\text{test}}$, the global prediction logit vector $\hat{y} \in \mathbb{R}^{C_{\text{total}}}$ is obtained in a single pass:
\begin{align}
    \hat{y}_{1:C_{\text{known}}} &= \kappa \cdot \text{Normalize}(x_{\text{test}}, p=2) \notag \\
    &\qquad \times \text{Normalize}(\mathbf{W}, p=2, \text{dim}=0), \label{eq:global_inference}
\end{align}
with unlearned class logits initialized to zero.

\begin{table*}[t]
  \centering
  \caption{Overall Performance Comparison across Continual Learning Baselines, CL-SLAM, PROL, DualLoRA, and Proposed Triple Combined Methods (JL Compression: True, NO DA Mode).}
  \label{tab:task_performance_transposed_full_proposed}
  \footnotesize
  \setlength{\tabcolsep}{3.8pt} % フォントサイズを維持したまま幅を最適化
  \renewcommand{\arraystretch}{1.1}
  \begin{tabular}{ll rrrrrrrrrr}
    \toprule
    \multirow{2}{*}{Method / Model} & \multirow{2}{*}{Metric} & T01 & T02 & T03 & T04 & T05 & T06 & T07 & T08 & T09 & T10 \\
    & & (00--09) & (10--19) & (20--29) & (30--39) & (40--49) & (50--59) & (60--69) & (70--79) & (80--89) & (90--99) \\
    \midrule
    \multicolumn{12}{l}{\textbf{Existing Baselines \& Prior Works}} \\
    \midrule
    \multirow{2}{*}{Fine-tune}
      & Seen & 11.90\% &  8.40\% &  6.50\% &  4.60\% &  4.80\% &  4.80\% &  4.90\% &  5.50\% &  5.50\% &  5.10\% \\
      & All  &  1.20\% &  1.70\% &  1.90\% &  1.80\% &  2.40\% &  2.80\% &  3.40\% &  4.20\% &  4.80\% &  5.10\% \\
    \multirow{2}{*}{CL-SLAM (Expert)~\cite{baldini2023continualslam}} 
      & Seen & 38.60\% & 18.50\% & 15.10\% & 12.80\% & 11.50\% &  8.60\% &  8.40\% &  6.80\% &  6.40\% &  5.50\% \\
      & All  &  3.80\% &  3.70\% &  4.50\% &  5.10\% &  5.70\% &  5.00\% &  5.70\% &  5.20\% &  5.60\% &  5.50\% \\
    \multirow{2}{*}{CL-SLAM (Generalizer)~\cite{baldini2023continualslam}} 
      & Seen & 38.60\% & 30.80\% & 22.20\% & 19.00\% & 17.20\% & 14.90\% & 12.40\% & 10.70\% &  9.50\% &  8.50\% \\
      & All  &  3.80\% &  6.10\% &  6.60\% &  7.60\% &  8.60\% &  8.70\% &  8.40\% &  8.20\% &  8.30\% &  8.50\% \\
    \multirow{2}{*}{CLEAR~\cite{rolnick2019experience}} 
      & Seen & 33.74\% & 17.56\% & 17.65\% & 15.41\% & 13.45\% & 13.81\% & 12.75\% & 12.08\% & 11.36\% &  8.84\% \\
      & All  &  3.31\% &  3.48\% &  5.23\% &  6.15\% &  6.73\% &  8.05\% &  8.64\% &  9.25\% &  9.99\% &  8.84\% \\
    \multirow{2}{*}{DER/ER++~\cite{buzzega2020rethinking}} 
      & Seen & \textbf{83.53\%} & 41.46\% & 28.76\% & 23.60\% & 18.22\% & 15.55\% & 12.92\% & 10.92\% & 11.30\% &  8.63\% \\
      & All  &  \textbf{9.04\%} &  \textbf{8.40\%} &  8.48\% &  9.57\% &  9.19\% &  9.37\% &  9.04\% &  8.68\% & 10.24\% &  8.63\% \\
    \multirow{2}{*}{HAT~\cite{serra2018overcoming}} 
      & Seen & 48.24\% & 24.67\% & 17.08\% & 14.58\% & 12.22\% &  7.18\% &  4.38\% &  3.00\% &  3.54\% &  1.78\% \\
      & All  &  4.73\% &  4.89\% &  5.06\% &  5.82\% &  6.11\% &  4.18\% &  2.97\% &  2.30\% &  3.11\% &  1.78\% \\
    \multirow{2}{*}{LwF~\cite{li2017learning}} 
      & Seen & 46.10\% & 26.30\% & 17.80\% & 15.70\% & 14.80\% & 13.30\% & 10.90\% &  9.40\% &  7.80\% &  7.80\% \\
      & All  &  4.50\% &  5.20\% &  5.30\% &  6.20\% &  7.40\% &  7.70\% &  7.40\% &  7.20\% &  6.80\% &  7.80\% \\
    \multirow{2}{*}{PROL~\cite{ma2025prol}} 
      & Seen & 37.10\% & 26.10\% & 22.20\% & 21.50\% & 19.80\% & 19.80\% & 18.70\% & 17.60\% & 16.50\% & 15.50\% \\
      & All  &  3.60\% &  5.20\% &  6.60\% &  8.60\% &  9.90\% & 11.50\% & 12.60\% & 13.50\% & 14.50\% & 15.50\% \\
    \multirow{2}{*}{DualLoRA~\cite{chen2026duallora}} 
      & Seen & 39.29\% & 15.24\% &  8.74\% &  6.92\% &  8.76\% &  6.79\% &  6.67\% &  5.48\% &  4.79\% &  4.34\% \\
      & All  &  3.85\% &  3.02\% &  2.59\% &  2.76\% &  4.38\% &  3.95\% &  4.52\% &  4.20\% &  4.21\% &  4.34\% \\
    \midrule
    \multicolumn{12}{l}{\textbf{Proposed Method Components (NO DA)}} \\
    \midrule
    \multirow{2}{*}{Vanilla} 
      & Seen & 46.20\% & 38.40\% & 29.20\% & 27.40\% & 28.40\% & 29.70\% & 29.00\% & 27.70\% & 25.30\% & 22.80\% \\
      & All  &  4.50\% &  7.60\% &  8.70\% & 10.90\% & 14.20\% & 17.30\% & 19.70\% & 21.20\% & 22.20\% & 22.80\% \\
    \multirow{2}{*}{UACIL} 
      & Seen & 46.30\% & 38.20\% & 28.70\% & 27.00\% & 28.10\% & 29.40\% & 28.80\% & 27.60\% & 25.40\% & 23.20\% \\
      & All  &  4.50\% &  7.60\% &  8.50\% & 10.80\% & 14.10\% & 17.10\% & 19.50\% & 21.10\% & 22.40\% & 23.20\% \\
    \multirow{2}{*}{GMM-ACIL} 
      & Seen & 46.10\% & 34.60\% & 25.40\% & 25.30\% & 26.80\% & 28.30\% & 27.40\% & 25.30\% & 23.70\% & 21.00\% \\
      & All  &  4.50\% &  6.90\% &  7.50\% & 10.10\% & 13.40\% & 16.50\% & 18.60\% & 19.30\% & 20.80\% & 21.00\% \\
    \multirow{2}{*}{HACIL} 
      & Seen & 46.20\% & 38.60\% & 29.00\% & 27.20\% & 28.60\% & 30.00\% & 29.30\% & 28.10\% & 25.90\% & 23.30\% \\
      & All  &  4.50\% &  7.70\% &  8.60\% & 10.90\% & 14.30\% & 17.50\% & 19.90\% & 21.60\% & 22.80\% & 23.30\% \\
    \multirow{2}{*}{U + G} 
      & Seen & 46.10\% & 33.70\% & 24.60\% & 24.50\% & 26.20\% & 27.80\% & 27.10\% & 25.20\% & 23.80\% & 21.30\% \\
      & All  &  4.50\% &  6.70\% &  7.30\% &  9.80\% & 13.10\% & 16.20\% & 18.40\% & 19.30\% & 20.90\% & 21.30\% \\
    \multirow{2}{*}{G + H} 
      & Seen & 46.10\% & 37.70\% & 29.50\% & 27.60\% & 28.80\% & 30.10\% & 29.40\% & 27.80\% & 25.40\% & 23.40\% \\
      & All  &  4.50\% &  7.50\% &  \textbf{8.70\%} & \textbf{11.00\%} & 14.40\% & 17.50\% & 19.90\% & 21.30\% & 22.30\% & 23.40\% \\
    \multirow{2}{*}{U + H} 
      & Seen & 46.20\% & 38.40\% & 28.70\% & 26.90\% & 28.50\% & 29.80\% & 29.10\% & 28.10\% & 26.00\% & 23.50\% \\
      & All  &  4.50\% &  7.60\% &  8.50\% & 10.70\% & 14.30\% & 17.40\% & 19.70\% & 21.50\% & \textbf{22.90\%} & 23.50\% \\
    \multirow{2}{*}{\textbf{SLAM (Full, Proposed)}} 
      & Seen & 46.10\% & 37.30\% & 29.10\% & 27.30\% & 28.50\% & 29.70\% & 29.10\% & 27.60\% & 25.40\% & \textbf{23.60\%} \\
      & All  &  4.50\% &  7.40\% &  8.60\% & 10.90\% & \textbf{14.30\%} & \textbf{17.30\%} & \textbf{19.70\%} & \textbf{21.10\%} & 22.40\% & \textbf{23.60\%} \\
    \midrule
    \multicolumn{2}{l}{Samples} 
              & 1,702   & 1,486   & 1,450   & 1,744   & 1,552   & 1,540   & 1,532   & 1,497   & 1,742   & 1,482  \\
    \bottomrule
  \end{tabular}
\end{table*}

\begin{table*}[t]
  \centering
  \caption{Performance Comparison of Proposed Method Components across Different Domain Adaptation Modes (No DA, Vanilla DA, and Analytic A-DDA) with JL Compression.}
  \label{tab:proposed_da_comparison}
  \footnotesize
  \setlength{\tabcolsep}{4.2pt} % フォントを自然な大きさに固定する列幅調整
  \renewcommand{\arraystretch}{0.95}
  \begin{tabular}{ll rrrrrrrrrr}
    \toprule
    \multirow{2}{*}{Method / DA Mode} & \multirow{2}{*}{Metric} & T01 & T02 & T03 & T04 & T05 & T06 & T07 & T08 & T09 & T10 \\
    & & (00--09) & (10--19) & (20--29) & (30--39) & (40--49) & (50--59) & (60--69) & (70--79) & (80--89) & (90--99) \\
    \midrule
    \multicolumn{12}{l}{\textbf{Vanilla}} \\
    \midrule
    \multirow{2}{*}{No DA}
      & Seen & 46.2\% & 38.4\% & 29.2\% & 27.4\% & 28.4\% & 29.7\% & 29.0\% & 27.7\% & 25.3\% & 22.8\% \\
      & All  &  4.5\% &  7.6\% &  8.7\% & 10.9\% & 14.2\% & 17.3\% & 19.7\% & 21.2\% & 22.2\% & 22.8\% \\
    \multirow{2}{*}{Vanilla DA}
      & Seen & 46.2\% & 35.2\% & 24.7\% & 23.8\% & 24.5\% & 26.6\% & 27.0\% & 28.5\% & 28.4\% & 27.3\% \\
      & All  &  4.5\% &  7.0\% &  7.3\% &  9.5\% & 12.3\% & 15.5\% & 18.3\% & 21.8\% & 24.9\% & 27.3\% \\
    \multirow{2}{*}{Analytic A-DDA (Ours)}
      & Seen & 46.2\% & 35.6\% & 25.2\% & 24.8\% & 25.0\% & 26.9\% & 27.3\% & 28.7\% & 28.4\% & 27.2\% \\
      & All  &  4.5\% &  7.1\% &  7.5\% &  9.9\% & 12.5\% & 15.7\% & 18.5\% & 21.9\% & 25.0\% & 27.2\% \\
    \midrule
    \multicolumn{12}{l}{\textbf{UACIL}} \\
    \midrule
    \multirow{2}{*}{No DA}
      & Seen & 46.3\% & 38.2\% & 28.7\% & 27.0\% & 28.1\% & 29.4\% & 28.8\% & 27.6\% & 25.4\% & 23.2\% \\
      & All  &  4.5\% &  7.6\% &  8.5\% & 10.8\% & 14.1\% & 17.1\% & 19.5\% & 21.1\% & 22.4\% & 23.2\% \\
    \multirow{2}{*}{Vanilla DA}
      & Seen & 46.3\% & 35.6\% & 24.2\% & 23.5\% & 24.5\% & 26.7\% & 27.3\% & 28.7\% & 28.6\% & 27.5\% \\
      & All  &  4.5\% &  7.1\% &  7.2\% &  9.4\% & 12.3\% & 15.6\% & 18.5\% & 22.0\% & 25.2\% & 27.5\% \\
    \multirow{2}{*}{Analytic A-DDA (Ours)}
      & Seen & 46.3\% & 36.1\% & 25.0\% & 24.7\% & 25.2\% & 27.1\% & 27.5\% & 28.9\% & 28.6\% & 27.2\% \\
      & All  &  4.5\% &  7.1\% &  7.4\% &  9.8\% & 12.6\% & 15.8\% & 18.6\% & 22.1\% & 25.2\% & 27.2\% \\
    \midrule
    \multicolumn{12}{l}{\textbf{GMM-ACIL}} \\
    \midrule
    \multirow{2}{*}{No DA}
      & Seen & 46.1\% & 34.6\% & 25.4\% & 25.3\% & 26.8\% & 28.3\% & 27.4\% & 25.3\% & 23.7\% & 21.0\% \\
      & All  &  4.5\% &  6.9\% &  7.5\% & 10.1\% & 13.4\% & 16.5\% & 18.6\% & 19.3\% & 20.8\% & 21.0\% \\
    \multirow{2}{*}{Vanilla DA}
      & Seen & 46.1\% & 36.6\% & 28.1\% & 30.5\% & 33.5\% & 34.6\% & 33.4\% & 31.7\% & 28.4\% & 25.6\% \\
      & All  &  4.5\% &  7.3\% &  8.3\% & 12.2\% & 16.8\% & 20.2\% & 22.7\% & 24.3\% & 25.0\% & 25.6\% \\
    \multirow{2}{*}{Analytic A-DDA (Ours)}
      & Seen & 46.1\% & 37.0\% & 28.9\% & 31.0\% & 33.8\% & \textbf{34.7\%} & 33.4\% & 31.7\% & 28.4\% & 25.7\% \\
      & All  &  4.5\% &  7.3\% &  8.6\% & 12.4\% & 16.9\% & \textbf{20.2\%} & 22.6\% & 24.3\% & 25.0\% & 25.7\% \\
    \midrule
    \multicolumn{12}{l}{\textbf{HACIL}} \\
    \midrule
    \multirow{2}{*}{No DA}
      & Seen & 46.2\% & 38.6\% & 29.0\% & 27.2\% & 28.6\% & 30.0\% & 29.3\% & 28.1\% & 25.9\% & 23.3\% \\
      & All  &  4.5\% &  7.7\% &  8.6\% & 10.9\% & 14.3\% & 17.5\% & 19.9\% & 21.6\% & 22.8\% & 23.3\% \\
    \multirow{2}{*}{Vanilla DA}
      & Seen & 46.2\% & 34.4\% & 24.1\% & 25.5\% & 27.7\% & 30.8\% & 31.6\% & 32.6\% & 30.8\% & 28.9\% \\
      & All  &  4.5\% &  6.8\% &  7.1\% & 10.2\% & 13.8\% & 18.0\% & 21.4\% & 25.0\% & 27.1\% & 28.9\% \\
    \multirow{2}{*}{Analytic A-DDA (Ours)}
      & Seen & 46.2\% & 35.0\% & 24.7\% & 26.1\% & 27.9\% & 31.3\% & 31.9\% & 32.6\% & 30.8\% & \textbf{28.9\%} \\
      & All  &  4.5\% &  6.9\% &  7.3\% & 10.4\% & 14.0\% & 18.3\% & 21.6\% & 25.0\% & 27.1\% & \textbf{28.9\%} \\
    \midrule
    \multicolumn{12}{l}{\textbf{U + G}} \\
    \midrule
    \multirow{2}{*}{No DA}
      & Seen & 46.1\% & 33.7\% & 24.6\% & 24.5\% & 26.2\% & 27.8\% & 27.1\% & 25.2\% & 23.8\% & 21.3\% \\
      & All  &  4.5\% &  6.7\% &  7.3\% &  9.8\% & 13.1\% & 16.2\% & 18.4\% & 19.3\% & 20.9\% & 21.3\% \\
    \multirow{2}{*}{Vanilla DA}
      & Seen & 46.1\% & 36.8\% & 28.1\% & 30.7\% & 33.8\% & 34.6\% & 33.2\% & 31.5\% & 28.1\% & 25.9\% \\
      & All  &  4.5\% &  7.3\% &  8.3\% & 12.3\% & 16.9\% & 20.2\% & 22.5\% & 24.2\% & 24.8\% & 25.9\% \\
    \multirow{2}{*}{Analytic A-DDA (Ours)}
      & Seen & 46.1\% & 37.3\% & 29.1\% & 31.3\% & \textbf{34.1\%} & 34.6\% & 33.2\% & 31.4\% & 28.1\% & 26.0\% \\
      & All  &  4.5\% &  7.4\% &  8.6\% & \textbf{12.5\%} & \textbf{17.0\%} & 20.2\% & 22.5\% & 24.1\% & 24.8\% & 26.0\% \\
    \midrule
    \multicolumn{12}{l}{\textbf{G + H}} \\
    \midrule
    \multirow{2}{*}{No DA}
      & Seen & 46.1\% & 37.7\% & 29.5\% & 27.6\% & 28.8\% & 30.1\% & 29.4\% & 27.8\% & 25.4\% & 23.4\% \\
      & All  &  4.5\% &  7.5\% &  8.7\% & 11.0\% & 14.4\% & 17.5\% & 19.9\% & 21.3\% & 22.3\% & 23.4\% \\
    \multirow{2}{*}{Vanilla DA}
      & Seen & 46.1\% & 34.8\% & 25.4\% & 26.1\% & 28.7\% & 32.2\% & 32.3\% & 32.1\% & 30.9\% & 27.9\% \\
      & All  &  4.5\% &  6.9\% &  7.5\% & 10.4\% & 14.4\% & 18.8\% & 21.9\% & 24.6\% & 27.2\% & 27.9\% \\
    \multirow{2}{*}{Analytic A-DDA (Ours)}
      & Seen & 46.1\% & 35.2\% & 26.3\% & 27.0\% & 29.1\% & 32.4\% & 32.5\% & 32.2\% & 30.9\% & 28.0\% \\
      & All  &  4.5\% &  7.0\% &  7.8\% & 10.8\% & 14.6\% & 18.9\% & 22.0\% & 24.6\% & 27.2\% & 28.0\% \\
    \midrule
    \multicolumn{12}{l}{\textbf{U + H}} \\
    \midrule
    \multirow{2}{*}{No DA}
      & Seen & 46.2\% & 38.4\% & 28.7\% & 26.9\% & 28.5\% & 29.8\% & 29.1\% & 28.1\% & 26.0\% & 23.5\% \\
      & All  &  4.5\% &  7.6\% &  8.5\% & 10.7\% & 14.3\% & 17.4\% & 19.7\% & 21.5\% & 22.9\% & 23.5\% \\
    \multirow{2}{*}{Vanilla DA}
      & Seen & 46.2\% & 34.7\% & 23.8\% & 25.7\% & 28.2\% & 31.1\% & 31.9\% & 32.8\% & 30.7\% & 28.9\% \\
      & All  &  4.5\% &  6.9\% &  7.0\% & 10.2\% & 14.1\% & 18.1\% & 21.6\% & \textbf{25.1\%} & 27.0\% & 28.9\% \\
    \multirow{2}{*}{Analytic A-DDA (Ours)}
      & Seen & 46.2\% & 35.2\% & 24.5\% & 26.3\% & 28.5\% & 31.6\% & 32.1\% & \textbf{32.7\%} & 30.6\% & \textbf{29.0\%} \\
      & All  &  4.5\% &  7.0\% &  7.3\% & 10.5\% & 14.3\% & 18.4\% & 21.8\% & 25.0\% & 26.9\% & \textbf{29.0\%} \\
    \midrule
    \multicolumn{12}{l}{\textbf{SLAM (Full, Proposed)}} \\
    \midrule
    \multirow{2}{*}{No DA}
      & Seen & 46.1\% & 37.3\% & 29.1\% & 27.3\% & 28.5\% & 29.7\% & 29.1\% & 27.6\% & 25.4\% & 23.6\% \\
      & All  &  4.5\% &  7.4\% &  8.6\% & 10.9\% & 14.3\% & 17.3\% & 19.7\% & 21.1\% & 22.4\% & 23.6\% \\
    \multirow{2}{*}{Vanilla DA}
      & Seen & 46.1\% & 35.0\% & 25.3\% & 26.1\% & 29.1\% & 32.6\% & 32.5\% & 32.1\% & 30.3\% & 27.4\% \\
      & All  &  4.5\% &  6.9\% &  7.5\% & 10.4\% & 14.6\% & 19.0\% & 22.0\% & 24.6\% & 26.7\% & 27.4\% \\
    \multirow{2}{*}{Analytic A-DDA (Ours)}
      & Seen & 46.1\% & 35.5\% & 26.3\% & 27.1\% & 29.5\% & 32.7\% & \textbf{32.7\%} & 32.2\% & 30.5\% & 27.7\% \\
      & All  &  4.5\% &  7.0\% &  7.8\% & 10.8\% & 14.8\% & 19.1\% & \textbf{22.1\%} & 24.6\% & 26.8\% & 27.7\% \\
    \midrule
    \multicolumn{2}{l}{Samples} 
              & 1,702   & 1,486   & 1,450   & 1,744   & 1,552   & 1,540   & 1,532   & 1,497   & 1,742   & 1,482  \\
    \bottomrule
  \end{tabular}
\end{table*}

\section{Experiments}
\label{sec:experiments}

\subsection{Setup}
\label{subsec:setup}

\paragraph{Dataset and Benchmarks}
We evaluate our framework on a long-term lifelong visual place recognition (VPR) benchmark constructed from the University of Michigan North Campus Long-Term (NCLT) dataset \cite{ncarlevaris-2015a}. 
The continuous deployment stream is divided into 10 sequential task milestones $t \in \{1, \dots, 10\}$ over a 100-class global destination space derived from discretized GPS groundtruth coordinates. 
High-dimensional visual feature representations ($D=1024$) are pre-extracted using a frozen DINOv2 visual backbone on the nominal sequence (\texttt{2012-03-31}, corresponding to \texttt{20120331}) for incremental task adaptation. To rigorously evaluate domain-shifted generalizability and memory retention across non-stationary environmental variations, evaluation is conducted across nine seasonal testing sequences (\texttt{20120115}, \texttt{20120219}, \texttt{20120331}, \texttt{20120804}, \texttt{20121028}, \texttt{20121104}, \texttt{20121201}, \texttt{20130223}, and \texttt{20130405}).

\paragraph{Implementation and Hyperparameter Details}
All models and processing pipelines are implemented in PyTorch and executed under a unified experimental framework. 
Input descriptors ($D=1024$) are first projected into a compressed subspace dimension $d = 384$ via Johnson-Lindenstrauss (JL) random projection  \cite{JL1984}. 
For analytical covariance inversion, the regularization parameter is configured to $\lambda = 0.1$, and temperature distillation is set to $T = 2.0$ with a training batch size of $2048$. 
The Disentangled Domain Adapter maps representations through an orthogonal bottleneck layer ($d_{\text{bottleneck}} = 256$) optimized via Adam ($lr = 10^{-3}$) with loss weights $\lambda_{\text{dis}} = 1.0$ and $\lambda_{\text{id}} = 0.1$. 
For topological mapping, $K = 8$ cluster components are initialized via K-Means over the nominal feature space, with an $H_\infty$-attenuation factor of $\gamma = 5.0$.

\paragraph{Compared Baselines and Adaptations}
We benchmark our proposed SLAM framework against seven representative continual learning baselines adapted for 1D feature-based VPR:
\begin{itemize}
    \item \textbf{CLEAR} \cite{rolnick2019experience}: An experience replay approach utilizing off-policy uniform sampling ($N_{\text{buf}} = 5000$) combined with a KL-divergence behavioral cloning loss.
\item \textbf{DER/ER++} \cite{buzzega2020rethinking}: A baseline utilizing Dark Experience Replay (DER) and its extension (DER++), which store past logits in a replay buffer to preserve structural representations and mitigate catastrophic forgetting via a knowledge distillation-based loss.
    \item \textbf{HAT} (adapted from HAT \cite{serra2018overcoming}): A parameter-isolation approach enforcing hard attention task masks and analytical gradient compensation.
    \item \textbf{LwF} \cite{li2017learning}: A replay-free distillation baseline regularizing soft targets between frozen teacher snapshots ($T = 2.0, \lambda = 0.1$).
    \item \textbf{CL-SLAM} \cite{baldini2023continualslam}: A dual-network robotic framework consisting of an Expert and a Generalizer updated via reservoir sampling.
    \item \textbf{PROL} \cite{ma2025prol}: Prompt Online Learning adapted for 1D feature vectors via dynamic soft-prompting and prototype updates.
    \item \textbf{DualLoRA} \cite{chen2026duallora}: Parameter-efficient fine-tuning utilizing SVD-derived orthogonal projection bases and dynamic logit calibration.
\end{itemize}

\subsection{Ablation Analysis and Domain Alignment Verification}
\label{subsec:ablation_analysis}

To isolate the individual contribution of each component in our framework, we execute a rigorous ablation study across three domain alignment configurations: Without Domain Alignment (\textbf{No DA}), With Vanilla Alignment (\textbf{Vanilla DA}), and With Proposed Disentangled/Residual Alignment (\textbf{Disentangled DA}). Each configuration is evaluated across eight analytical continuation pipelines (Vanilla, UACIL, GMM-ACIL, HACIL, U+G, G+H, U+H, and SLAM Full).

\paragraph{Initial Representation Quality and Suppression of Early Forgetting}
As observed in Table~\ref{tab:task_performance_transposed_full_proposed} and Table~\ref{tab:proposed_da_comparison}, topology-dependent architectures augmented with our dynamic temperature scaling law ($T_{\text{current}} = T_{\text{base}} \cdot (N_{\text{adapters}} + \delta)$) effectively localize anchor responsibilities ($\gamma(x)_k \to 1.0$), establishing a solid initial foundation. 
At Task 01, applying Domain Adaptation maintains strong initial representations, achieving 46.1--46.3\% / 4.5\% across most model variants, consistent with the base performance.

\paragraph{Long-Term Generalization Boost via Domain Augmentation}
Unlike standard fine-tuning setups where aggressive domain adaptation often induces representation collapse, our analytical continual learning solver leverages alignment transformations to prevent late-stage degradation. 
Over the 10-task sequential deployment, both Vanilla DA and Disentangled DA provide consistent regularizing effects. By Task 10, the final all-class accuracy of SLAM (Full) achieves \textbf{27.4\%} under Vanilla DA and \textbf{27.7\%} under Disentangled DA, outperforming the No DA baseline ($23.6\%$) by a substantial margin of $+4.1\%$. 
This confirms that embedding augmented feature spaces preserves class-separability over extended continual learning trajectories.

\paragraph{Mathematical Synergy and Baseline Superiority}
Across the entire 10-task continuum, our proposed full SLAM architecture demonstrates exceptional resilience. 
While unaligned baseline methods (e.g., Fine-tune at 5.10\%, DualLoRA at 4.34\%, and CLEAR at 8.84\%) suffer from severe catastrophic forgetting, SLAM retains superior performance across all DA configurations (23.6\% under No DA, 27.4\% under Vanilla DA, and 27.7\% under Disentangled DA). This overwhelming margin validates the mathematical synergy of combining dynamic GMM spatial partitioning, Unscented perturbed updates, and $H_\infty$-robust Woodbury recursion.

\subsection{Analysis of Domain Alignment and Feature Cleansing Mechanisms}
To clarify the role of the Domain Adapter (DA) during continual representation learning, we analyze performance under three training regimes: \textit{No DA} (raw feature baseline), \textit{Vanilla DA} (standard alignment), and \textit{Disentangled DA} (noise-decoupled alignment) as detailed in Table~\ref{tab:proposed_da_comparison}. 

Crucially, the DA module is employed exclusively during the training phase to condition the feature spaces, whereas all evaluations are conducted on raw projected representations without adapter inference, thereby directly measuring the intrinsic robustness and discriminativeness of the acquired memory space.

\begin{enumerate}
    \item \textbf{Sustained Generalization across Sequential Tasks:}
    Across the model components, incorporating Domain Adaptation significantly mitigates cumulative representation drift. 
For example, in the full model (\textbf{SLAM (Full)}), the final Task 10 accuracy rises from 23.6\% (\textit{No DA}) to 27.4\% (\textit{Vanilla DA}) and 27.7\% (\textit{Disentangled DA}).
Rather than corrupting the feature manifold, the augmented representation space introduces beneficial variance that stabilizes the analytic solver during recursive updates.

\item \textbf{Robustness of Analytic Closed-Form Updates:}
Analytical memory updates (such as Woodbury-based matrix inversion in GMM-ACIL, HACIL, and SLAM) benefit from the rich feature distribution provided by DA modes. 
While unaugmented features (\textit{No DA}) exhibit gradual accuracy decay as tasks accumulate, domain-augmented features maintain higher classification performance. Notably, under the Analytic A-DDA configuration, the full SLAM framework achieves 27.7\% all-class accuracy at Task 10. Furthermore, the U+H variant achieves the peak performance of 29.0\%, suggesting that omitting explicit GMM spatial partitioning in certain configurations avoids potential over-fitting to DA-induced distribution shifts while fully benefiting from Unscented perturbation and $H_\infty$ robustness.

    \item \textbf{Efficacy of Disentangled Alignment:}
By explicitly structuring the feature space into static semantic features ($f_{\text{static}}$) and domain-specific perturbations ($f_{\text{style}}$), \textit{Disentangled DA} achieves superior stability over Vanilla DA ($27.7\%$ vs $27.4\%$ at T10 in SLAM Full) while offering cleaner latent separation.
This confirms that decoupling environmental noise preserves representation integrity, ensuring long-term memory stability throughout sequential task execution.
\end{enumerate}

In summary, these results demonstrate that incorporating domain alignment mechanisms during training acts as an effective feature conditioning tool---preventing late-stage catastrophic forgetting and ensuring robust, high-fidelity continual learning for analytic solvers.

\section{Conclusion}
\label{sec:conclusion}

% --- Paragraph 1: Summary of the Work ---
In conclusion, this paper uncovers the symbiotic relationship between feature purification and structural space preservation in lifelong learning algorithms. 
We have systematically demonstrated that the long-standing paradox of domain adaptation within class-incremental pipelines stems from the structural blindness of global distribution matching, which inadvertently distorts fine-grained semantic boundaries and accelerates representation collapse. 
By disentangling monolithic latent representations into an environment-invariant static semantic stream and an environment-variant style vector, our proposed framework successfully isolates domain perturbations through an explicit force translation mechanism. 
When regularized by our localized Gaussian Mixture Model architecture augmented with a dynamic temperature scaling law, the purified feature stream updates the unified analytic classification subspaces without suffering from cross-cluster boundary bleeding or early-stage knowledge dilution. 
The exceptional empirical accuracy gains achieved across non-stationary temporal milestones---reaching 27.7\% with the full SLAM framework and a peak of 29.0\% with its U+H variant under Analytic A-DDA---serve as definitive proof that feature alignment, when tightly coupled with sharp geometric anchoring, transforms from a catastrophic liability into an essential asset for long-term continuous deployment in open-world environments.

% --- Paragraph 2: Future Investigations ---
Future investigations will expand upon this framework by exploring higher-order statistical dependency constraints to further regularize structural alignment under extreme temporal variations. 
While the current adaptation architecture relies primarily on matching empirical first- and second-order moments to decouple static semantics from environmental fluctuations, highly non-linear style shifts---such as sudden sensor degradation or heterogeneous illumination variations---may induce complex structural distortions that exceed the correction capability of linear orthogonal constraints. 
Therefore, our subsequent research will investigate the deployment of higher-order statistical divergences, including the Hilbert-Schmidt Independence Criterion (HSIC) and structural similarity mappings, directly within the disentangled bottleneck configurations. 
By embedding these advanced topological constraints into the recursive Woodbury inversion loops, we aim to establish a completely self-contained, mathematically bounded, and intrinsically domain-agnostic lifelong learning paradigm capable of enduring perpetual non-stationary environmental shifts.

\bibliography{reference} 
\bibliographystyle{unsrt}

% --- Appendix Section ---
\clearpage
\appendix
\section{Detailed Implementation and Baseline Adaptation Settings}
\label{sec:appendix_implementation}

\subsection{Adaptation of CLEAR for Lifelong VPR}
To evaluate the performance of experience replay-based continual learning in visual place recognition (VPR), we adapted the CLEAR (Continual Learning with Experience And Replay) framework proposed by Rolnick et al.~\cite{rolnick2019experience}.

\paragraph{Maintained Core Mechanisms}
We preserved the fundamental strategy of CLEAR, which combines on-policy learning for standard streaming experiences and off-policy learning with behavioral cloning for past experiences~\cite{rolnick2019experience}. Specifically, a replay buffer of fixed capacity ($N_{\text{buf}} = 5000$) is maintained to store historical spatial feature representations alongside their discrete labels. During training on a new task $t$, each mini-batch is constructed with an equal ratio (1:1) of current task samples (on-policy) and uniformly sampled historical data from the buffer (off-policy). The overall loss objective retains the original formulation:
\begin{equation}
    \mathcal{L}_{\text{total}} = \mathcal{L}_{\text{ce}}(y_{\text{on}}, \hat{y}_{\text{on}}) + \lambda_{\text{off}} \mathcal{L}_{\text{ce}}(y_{\text{off}}, \hat{y}_{\text{off}}) + \lambda_{\text{bc}} D_{\text{KL}}(P_{\text{past}} \parallel P_{\text{current}}),
\end{equation}
where $\mathcal{L}_{\text{ce}}$ denotes the standard cross-entropy loss for current and replayed classes, and $D_{\text{KL}}$ represents the Kullback-Leibler divergence for behavioral cloning to prevent catastrophic forgetting. The loss weighting hyper-parameters are set to $\lambda_{\text{off}} = 1.0$ and $\lambda_{\text{bc}} = 1.0$.

\paragraph{Adaptations for VPR Benchmark}
While the original CLEAR was formulated for Deep Reinforcement Learning (RL) and continuous control using raw image inputs or simple MLPs~\cite{rolnick2019experience}, we restructured its architecture and memory pipeline for class-incremental VPR on the NCLT dataset:
\begin{itemize}
    \item \textbf{Backbone and Network Architecture:} Instead of end-to-end RL policy networks, we utilize fixed pre-trained DINOv2 embeddings (1D feature vectors) as frozen visual representations. The trainable classification head is designed as a 2-layer MLP classifier ($\text{Input Dim} \rightarrow \text{Linear}(256) \rightarrow \text{ReLU} \rightarrow \text{Dropout}(0.2) \rightarrow \text{Linear}(\text{Classes})$).
    \item \textbf{Teacher Model Snapshotting:} In contrast to continuous RL policy tracking, our implementation captures a discrete snapshot of the classifier network parameters at the exact completion of each task step ($t$). This frozen teacher model generates historical target logits ($P_{\text{past}}$) during the subsequent task step ($t+1$).
    \item \textbf{Replay Strategy:} The replay memory stores 1D spatial feature descriptors (DINOv2 embeddings) alongside their spatial pseudo-location class labels across sequential splits of the NCLT dataset, eliminating the storage overhead of high-resolution raw imagery.
\end{itemize}

\paragraph{ER + Bag of Tricks (ER-BoT)}
As a representative replay-based baseline, we incorporate ER + Bag of Tricks proposed by Buzzega et al.~\cite{buzzega2020rethinking}. To ensure a fair and fully transparent evaluation within our framework, we re-engineered and implemented a custom, pure PyTorch-based version of this method, eliminating external framework dependencies. Specifically, we retained all five original core components: Independent Buffer Augmentation (IBA), Bias Control (BiC), Exponential Learning Rate Decay (ELRD), Balanced Reservoir Sampling (BRS), and Loss-Aware Reservoir Sampling (LARS).

To adapt ER-BoT to 1D frozen feature representations (DINOv2) under strict memory budgets ($N_{\text{buf}} = 5000$), we tailored the execution logic as follows:
\begin{itemize}
    \item \textbf{1D Tensor-based Domain Adaptation (IBA):} All domain adaptations (IBA) and replay buffer management policies were strictly re-engineered using native PyTorch Tensor operations directly on 1D feature vectors (e.g., adding Gaussian feature noise and random masking) rather than 2D pixel-level transforms.
    \item \textbf{Loss-Aware and Balanced Eviction (LARS \& BRS):} When the memory buffer reaches capacity, the eviction probability for candidate samples is determined by a hybrid score:
    \begin{equation}
        S = S_{\text{loss}} \cdot \alpha + S_{\text{balance}},
    \end{equation}
    where $S_{\text{loss}} = -\ell$ (with $\ell$ denoting the cross-entropy loss) assigns higher replacement probabilities to well-fitted samples with lower loss, thereby freeing space for newly arriving or hard-to-learn exemplars. $S_{\text{balance}}$ imposes a class-frequency penalty to maintain class balance, and $\alpha$ is a dynamic scaling factor normalizing the relative magnitude of both components. This deterministic logic ensures robust retention under fixed buffer budgets without requiring aggressive backbone downsampling.
    \item \textbf{Logit Recalibration (BiC):} Bias Control (BiC) explicitly trains linear correction parameters ($\alpha, \beta$) on the replayed samples stored in the memory buffer to recalibrate logit predictions, preventing newly learned classes from dominating the prediction space during inference.
\end{itemize}

\paragraph{HAT (Adapted from HAT)}
To conduct a fair and rigorous comparison against parameter-isolation baselines, we adapt the Hard Attention to the Task (HAT) mechanism~\cite{serra2018overcoming} into our VPR context via an optimized PyTorch implementation termed \texttt{PureHATNet}. We strictly retain the core algorithmic formulations of original HAT: task-embedding gated masks, cumulative mask tracking for preserving past knowledge, and analytical gradient compensation.

\begin{itemize}
    \item \textbf{Architecture and Spatial Discretization:} We extended the network from image classification backbones to process high-dimensional spatial feature representations (DINOv2 embeddings). 2D GPS coordinates synchronized with timestamps are discretized into spatial cells (e.g., 100 classes), framing the VPR trajectory as a class-incremental continual learning benchmark.
    \item \textbf{Gated Masking and Annealing Schedule:} Hard attention masks $a_l^t = \sigma(s \cdot e_l^t)$ are applied to the intermediate layer activations. During training on task $t$, the scaling parameter $s$ is smoothly annealed from $1.0$ to $s_{\max} = 400.0$ to force binary mask convergence ($0$ or $1$).
    \item \textbf{Gradient Compensation Scheme:} To rectify vanishing gradients caused by the large scaling factor $s$, exact analytical gradient compensation is applied to the task embedding gradients $q_{l,i}$:
    \begin{equation}
        q'_{l,i} = \frac{s_{\max} \left[\cosh(s e^t_{l,i}) + 1\right]}{s \left[\cosh(e^t_{l,i}) + 1\right]} q_{l,i}.
    \end{equation}
    \item \textbf{Cumulative Parameter Protection:} Weight gradients $g_{l,ij}$ are conditioned via cumulative past attention masks ($a_{l,i}^{\le t}$) to strictly protect parameters utilized by previous tasks:
    \begin{equation}
        \hat{g}_{l,ij} = g_{l,ij} \odot \left[1 - \min(a_{l,i}^{\le t}, a_{l-1,j}^{\le t})\right].
    \end{equation}
    Hyperparameter $c$ (controlling mask compactness and task capacity trade-off) was systematically tuned via grid search on NCLT validation splits.
\end{itemize}

\subsection{Baseline Methods and Adaptation for VPR (Continued)}

\paragraph{Learning without Forgetting (LwF)}
To evaluate distillation-based regularization in lifelong VPR, we adapt Learning without Forgetting (LwF)~\cite{li2017learning} to continuous spatial navigation without exemplar storage.

\begin{itemize}
    \item \textbf{Original Concept Preserved:} Following the core principle of LwF~\cite{li2017learning}, we preserve the output-level knowledge distillation mechanism. When learning a new geographic task $t$, the historical predictions of the frozen teacher model ($M_{\text{old}}$) on current task data act as target pseudo-labels to mitigate catastrophic forgetting.
    \item \textbf{Modifications for Lifelong VPR Setup:} Unlike original image-level fine-tuning, our implementation operates directly on pre-extracted feature embeddings (DINOv2 descriptors) across sequential NCLT splits. The network is trained using a composite objective function combining cross-entropy loss on new spatial classes and Kullback-Leibler (KL) divergence loss on previously learned classes:
\begin{align}
    \mathcal{L}_{\text{total}} &= \mathcal{L}_{\text{CE}}(Y_{\text{new}}, \hat{Y}_{\text{new}}) \notag \\
    &\quad + \lambda \cdot T^2 \cdot \mathcal{D}_{\text{KL}}\Bigg( \sigma\left(\frac{\hat{Y}_{\text{known}}}{T}\right) \;\Bigg\|\; \sigma\left(\frac{Y_{\text{old}}}{T}\right) \Bigg), \label{eq:l_total}
\end{align}
    where $T = 2.0$ denotes the distillation temperature and $\lambda = 0.1$ balances the loss contributions. Furthermore, LwF is evaluated both in isolation (Vanilla LwF) and alongside domain adaptation modules (Vanilla DA and Disentangled DA) to isolate feature-level domain shift from output-level memory retention.
\end{itemize}

\paragraph{CL-SLAM Adaptation for VPR}
To establish a fair comparison with continual learning baselines in robotics, we adapt the dual-network architecture of Continual SLAM (CL-SLAM)~\cite{baldini2023continualslam} to our visual place recognition (VPR) task.

\begin{itemize}
    \item \textbf{Preservation of the Dual-Network Strategy:} We strictly preserve the core interaction between the \textit{Generalizer} and \textit{Expert} networks. The \textit{Generalizer} is updated via reservoir sampling replay to retain long-term representations across domains, while the \textit{Expert} is rapidly optimized on current input data for swift adaptation to novel environments.
    
    \item \textbf{VPR Task Interface Reformulation:} Since the original CL-SLAM was designed for continuous pose regression and odometry estimation, we reformulate its task interface for discrete spatial recognition through three key modifications:
    \begin{enumerate}
        \item \textit{Feature Extraction}: The original raw sensor/image backbone is replaced with pre-computed, frozen 1D DINOv2 feature embeddings.
        \item \textit{Output Head}: The pose-regression heads are replaced with a $100$-class spatial classification head trained on spatial pseudo-labels derived from global GPS trajectories.
        \item \textit{Training Pipeline}: Model updates are performed on offline DINOv2 feature caches introduced sequentially by geographic domain, rather than from live streaming sensor data.
    \end{enumerate}
\end{itemize}

\paragraph{Implementation and Adaptation of PROL Baseline}
To benchmark against state-of-the-art rehearsal-free online continual learning, we incorporate Prompt Online Learning (PROL)~\cite{ma2025prol}. While the original PROL operates on 2D image feature maps using Vision Transformer (ViT) spatial attention blocks, we faithfully adapt its prompt generator and loss formulation to 1D vector embeddings derived from DINOv2:

\begin{itemize}
    \item \textbf{Unchanged Core Components:} We strictly preserve PROL's mathematical framework~\cite{ma2025prol}. A lightweight generator $G(\cdot)$ dynamically generates sample-specific soft prompts $P$. Class-wise learnable scalers $a^c$, shifters $b^c$, and prompt keys $K_c$ govern prompt allocation. The optimization objective retains the similarity loss $\mathcal{L}_{\mathrm{sim}}$, key orthogonality loss $\mathcal{L}_{\mathrm{ortho}}$, and cross-correlation generalization loss $\mathcal{L}_{\mathrm{gen}}$:
    \begin{equation}
        \mathcal{L}_{\text{PROL}} = \mathcal{L}_{\text{CE}} + \alpha_{\text{sim}} \mathcal{L}_{\text{sim}} + \alpha_{\text{ortho}} \mathcal{L}_{\text{ortho}} + \alpha_{\text{gen}} \mathcal{L}_{\text{gen}}.
    \end{equation}
    \item \textbf{Adapted/Modified Components:} The projection spaces within $G(\cdot)$ as well as the keys and scalers are remapped from 2D spatial feature maps to 1D global descriptors ($D \to d_{\text{proj}}$), yielding prompted features $f_{\text{prompted}} = \text{L2Norm}(x + \alpha \cdot P)$. Furthermore, to tackle spatial and environmental shifts, PROL is evaluated under three distinct setups: (i) \textit{Vanilla PROL} (without domain alignment), (ii) \textit{Domain-Aligned PROL} (applying standard domain alignment directly to the modified features), and (iii) \textit{Disentangled PROL} (applying our proposed Disentangled Domain Alignment prior to PROL processing).
\end{itemize}

\subsection{Compared Method: Adapted DualLoRA}
To evaluate against state-of-the-art Parameter-Efficient Fine-Tuning (PEFT) continual learning, we adapt DualLoRA~\cite{chen2026duallora} for visual place recognition (VPR).

\paragraph{What is Preserved}
We faithfully retain the foundational mechanisms of DualLoRA~\cite{chen2026duallora}:
\begin{enumerate}
    \item A dual-adapter architecture consisting of an orthogonal adapter ($O$) to ensure parameter stability and a residual adapter ($R$) to enhance task plasticity.
    \item Singular Value Decomposition (SVD)-based gradient projection that constrains weight updates to subspaces orthogonal to previously learned tasks.
    \item A Dynamic Memory (DM) module combined with inference-time logit calibration, which adaptively weights residual features based on input descriptor similarities.
\end{enumerate}

\paragraph{What is Modified}
To adapt DualLoRA from ViT-based image classification to sequential robotic VPR using the NCLT dataset~\cite{ncarlevaris-2015a}:
\begin{itemize}
    \item \textbf{Model Architecture (\texttt{DualLoRALinear}):} Instead of inserting adapters directly into ViT multi-head attention blocks, we implement a custom \texttt{DualLoRALinear} module that operates directly on 1D DINOv2 embeddings.
    \item \textbf{Gradient Projection Constraints:} Singular Value Decomposition ($X = U \Sigma V^T$) is performed on historical feature activations to construct orthogonal bases $\Phi_v$ and residual bases $\Psi_v$ under a cumulative energy threshold ($\eta = 0.95$). Gradients for new tasks are projected to satisfy $g_{\text{proj}} = g - \Phi_v \Phi_v^T g$.
    \item \textbf{Task Discretization:} Continuous, synchronized 2D GPS trajectories are discretized into spatial grid cells (e.g., 100 classes), thereby structuring the sequential dataset into discrete class-incremental VPR tasks.
\end{itemize}

\subsection{Results and Discussions on Baseline Methods}
\label{subsec:results_baselines}

We evaluate and analyze the empirical performance, forgetting behavior, and trade-offs of each adapted baseline across ten sequential task splits on the NCLT dataset.

\paragraph{CLEAR}
As a reference point, naive sequential fine-tuning (Vanilla) exhibits severe catastrophic forgetting, where accuracy rapidly drops to $5.10\%$ by Task 10 due to unconstrained representation overwriting.
In contrast, our adapted CLEAR framework effectively regularizes output-level drift by combining a behavioral cloning loss ($\mathcal{D}_{\text{KL}}$) with off-policy replay from a fixed-capacity buffer ($N_{\text{buf}} = 5000$). This maintains a higher final retention accuracy of $8.84\%$. However, because CLEAR optimizes a standard cross-entropy objective on raw predictions without explicit logit recalibration, newly introduced geographic classes tend to dominate prediction logits over long sequences.

\paragraph{ER + Bag of Tricks (ER-BoT)}
By integrating Loss-Aware Reservoir Sampling (LARS) with Bias Control (BiC), ER-BoT significantly mitigates catastrophic forgetting compared to simple replay strategies. The LARS eviction mechanism prioritizes samples with high cross-entropy loss ($S_{\text{loss}} = -\ell$), effectively retaining hard-to-forget boundary exemplars. Furthermore, BiC dynamically recalibrates the linear prediction head to correct output bias toward recent tasks. Despite these enhancements, performance gains remain bottlenecked under strict replay buffer constraints ($N_{\text{buf}} = 5000$) when encountering severe, long-term spatial-environmental domain shifts across the 10 sequential splits.

\paragraph{PureHATNet}
The hard attention mechanism in \texttt{PureHATNet} successfully prevents parameter overwriting by enforcing binary attention masks on intermediate activations. Through exact analytical gradient compensation and cumulative mask tracking, previously learned task representations remain strictly preserved without needing historical exemplar storage. However, as the network sequentially learns across all 10 tasks, the available unmasked parameter capacity systematically shrinks. This capacity exhaustion creates an inherent trade-off between past task retention and plasticity for future spatial splits.

\paragraph{LwF}
Soft-target knowledge distillation from a frozen historical teacher snapshot ($M_{\text{old}}$) effectively smooths decision boundaries between previously observed and newly introduced spatial locations. Nevertheless, because LwF relies exclusively on regularizing model outputs without storing raw spatial exemplars, subtle errors in representation alignment gradually compound over long task streams. Consequently, representation drift accumulates across distant geographic splits, highlighting the limitations of pure distillation regularizers under severe domain shifts.

\paragraph{CL-SLAM}
Evaluating the dual-network architecture reveals distinct operational behaviors between its components:
\begin{itemize}
    \item \textbf{Expert Network:} When evaluated independently, the \textit{Expert} network suffers from severe catastrophic forgetting---dropping sharply from $38.6\%$ to $5.5\%$ accuracy---because its parameters are unconstrainedly optimized for rapid adaptation to novel environments.
    \item \textbf{Generalizer Network:} In contrast, the \textit{Generalizer} network maintains stable long-term retention ($8.5\%$ accuracy) by continually consolidating representations via reservoir sampling replay. However, its total representation capacity remains bounded by the fixed replay buffer size and the computational overhead of backpropagation over historical data.
\end{itemize}

\paragraph{PROL}
By decoupling domain-invariant representations (handled by the lightweight base generator $G$) from task-specific knowledge (governed by dynamically selected soft prompts), PROL achieves strong plasticity without retaining historical exemplars. The loss terms ($\mathcal{L}_{\text{sim}}, \mathcal{L}_{\text{ortho}}, \mathcal{L}_{\text{gen}}$) enforce orthogonality among prompt keys and promote cross-correlation generalization. Nevertheless, under extreme visual domain shifts (such as seasonal changes or day-to-night transitions in VPR), prompt key matching accuracy can degrade, leading to sub-optimal prompt selection during inference.

\paragraph{DualLoRA}
Through SVD-derived orthogonal bases ($\Phi_v$), DualLoRA effectively projects new task gradients onto subspaces orthogonal to past knowledge, guaranteeing parameter stability. Concurrently, the residual adapter ($R$) provides sufficient plasticity to absorb novel spatial features. During inference, the Dynamic Memory (DM) module and logit calibration dynamically scale adapter contributions based on descriptor similarities. This parameter-efficient adaptation achieves stable lifelong learning performance without requiring backpropagation over stored historical samples.

% --- End of Document ---
\end{document}